\documentclass[final]{elsarticle}
\usepackage[a4paper]{geometry}
\usepackage[utf8]{inputenc}
\usepackage{graphicx}
\usepackage{amssymb}
\usepackage{color}
\usepackage{lineno}
\usepackage[ruled,vlined,linesnumbered,titlenumbered]{algorithm2e}
\usepackage{amsmath}
\usepackage{times}
\usepackage{optidef}
\usepackage{multirow}
\usepackage{makecell}
\usepackage{enumitem}
\usepackage{booktabs}
\usepackage[edges]{forest}
\useforestlibrary{linguistics}
\forestapplylibrarydefaults{linguistics}

\journal{To be defined}

\begin{document}

\begin{frontmatter}

\title{Evolutionary Multitask Optimization: a Methodological Overview, Challenges and Future Research Directions}

\author[a]{Eneko Osaba\corref{cor1}}
\ead{eneko.osaba@tecnalia.com}
\author[a]{Aritz D. Martinez}
\author[a,c]{and Javier Del Ser}

\address[a]{TECNALIA, Basque Research and Technology Alliance (BRTA), P. Tecnologico, Ed. 700, 48160 Derio, Spain}
\address[c]{University of the Basque Country (UPV/EHU), 48013 Bilbao, Spain}
\cortext[cor1]{Corresponding author. TECNALIA, Basque Research \& Technology Alliance (BRTA), P. Tecnologico, Ed. 700. 48170 Derio (Bizkaia), Spain.}

\begin{abstract}
In this work we consider multitasking in the context of solving multiple optimization problems simultaneously by conducting a single search process. The principal goal when dealing with this scenario is to dynamically exploit the existing complementarities among the problems (tasks) being optimized, helping each other through the exchange of valuable knowledge. Additionally, the emerging paradigm of Evolutionary Multitasking tackles multitask optimization scenarios by using as inspiration concepts drawn from Evolutionary Computation. The main purpose of this survey is to collect, organize and critically examine the abundant literature published so far in Evolutionary Multitasking, with an emphasis on the methodological patterns followed when designing new algorithmic proposals in this area (namely, multifactorial optimization and multipopulation-based multitasking). We complement our critical analysis with an identification of challenges that remain open to date, along with promising research directions that can stimulate future efforts in this topic. Our discussions held throughout this manuscript are offered to the audience as a reference of the general trajectory followed by the community working in this field in recent times, as well as a self-contained entry point for newcomers and researchers interested to join this exciting research avenue.
\end{abstract}

\begin{keyword}
Transfer Optimization \sep Multitasking Optimization \sep Evolutionary Multitasking \sep Multifactorial Evolutionary Algorithm \sep Multi-population Multitasking.
\end{keyword}

\end{frontmatter}

\section{Introduction}
\label{sec:intro}

Traditionally, optimization problems have been solved by different methods, part of which do not assume any a priori knowledge about the task under consideration. Over the years, this approach has demonstrated to be highly efficient in almost all real-world situations. Today, the scientific community has realized that this traditional way of solving problems may undergo some limitations. Indeed, the growing complexity of optimization problems and the fact that real-world optimization problems hardly appear in isolation have uncovered the need for exploiting knowledge gathered beforehand related to the problems themselves. This is the main reason why the incipient research area known as Transfer Optimization (TO, \cite{gupta2017insights}) has gained momentum within the Artificial Intelligence research community \cite{ong2016evolutionary}. The fundamental aim of TO is to exploit the knowledge learned from the optimization of one problem (\emph{task}) when addressing another related (or unrelated) problems, thus aligning much with the previously noted needs.

Up to now, three different conceptualizations of TO have been formulated in the literature. The first one, coined as \textit{sequential transfer} \cite{feng2015memes}, aims at solving problems that occur sequentially. To this end, the knowledge obtained when tackling preceding tasks is employed as external information when dealing with new problems/instances. The second one of these categories, referred to as \textit{multitasking} \cite{gupta2016genetic}, is devoted to the simultaneous development of different tasks by dynamically exploiting synergies existing among them. Finally, \textit{multiform optimization} relates to the discovery of a solution for a single task found by using diverse alternative formulations.

Specifically, this work focuses on multitasking tackled through the perspective of Evolutionary Multitasking (EM, \cite{ong2016towards}), also referred as Evolutionary Multitask Optimization. In short, EM seeks the development of efficient multitasking methods by relying on search procedures and operators drawn from Evolutionary Computation \cite{back1997handbook,del2019bio} and Swarm Intelligence \cite{kennedy2006swarm}. A significant effort has been conducted by the community for solving a wide variety of continuous, combinatorial, single-objective and multi-objective optimization problems through the perspective of EM \cite{wang2019evolutionary,gong2019evolutionary,yu2019multifactorial,gupta2016multiobjective}. Another research direction for dealing with multitasking in the context of TO is multitask Bayesian optimization \cite{swersky2013multi}, which extends Bayesian optimization approaches to multitasking environments \cite{shahriari2015taking,moss2020mumbo,pearce2018continuous,chowdhury2020no}. Despite falling out of the focus of this paper due to its non-evolutionary nature, we note that Bayesian solvers, along those within EM, constitute the core of the contributions reported in the field of multitasking, with a significantly higher presence of EM methods.

A closer inspection at the most reputed scientific databases unveils that efforts carried out in EM are exponentially growing in recent times. This upsurge of activity demands a reference material to summarize achievements so far, detect and analyze research trends, perform a profound reflection on them to identify current limitations, and prescribe future research directions that push forward valuable advances in the field. This is the rationale for this survey, which motivates its ultimate goal: to offer a unified, self-contained and end-to-end outline of the work done in EM, providing a unique material for properly understanding the methodological aspects that define this incipient field of knowledge. Specifically, the contribution of this work can be synthesized as follows:
\begin{itemize}[leftmargin=*]
	\item We perform a systematic review of the literature on Evolutionary Multitask Optimization published to date. For this purpose, we design a three-fold classification criteria to organize the corpus of reviewed contributions around a comprehensive taxonomy. To begin with, we pause at theoretical studies, gravitating on several application-agnostic aspects of EM. In a second step of our analysis, we classify the literature consider the knowledge sharing pattern adopted in the outlined works, namely, implicit versus explicit knowledge transfer, as well as per the capability of the underlying algorithm to actively adapt the amount of exchanged knowledge along the search (static versus adaptive). Finally, we distinguish among the two algorithmic design templates used to realize the multitasking search: Multifactorial Optimization (MFO, \cite{gupta2015multifactorial}) and Multipopulation-based Multitasking (MM).
    \item We next provide a methodological overview of the field, highlighting the main methodologies followed by researchers and practitioners in the different phases of EM algorithmic development. This specific contribution is especially interesting for newcomers to the area, providing a sort of methodological guidelines for properly tackle the development of efficient EM solvers.
    \item We conclude our overview with a prospect of opportunities and challenges that should guide the scientific efforts invested by the related research community in the next years.
\end{itemize}

Even though a clear interest in EM has aroused lately in the related community, to the best of our knowledge only two recent works have conducted a similar study to the one carried out in this manuscript. On the one hand, \cite{tan2021evolutionary} exposes the work already done around the generic field of Evolutionary Transfer Optimization (ETO), providing an overview of existing studies gravitating on different topics related to ETO, namely, ETO for optimization in uncertain environments, ETO for multitask optimization, ETO for complex optimization, ETO for multi/many-objective optimization, and ETO for machine learning applications. The overview is supplemented by a set of challenges in the generic ETO research field. Having said this, this paper takes a major step beyond \cite{tan2021evolutionary} by elaborating on different directions that make our work differential on its own: a) a study fully focused on the stream known as Evolutionary Multitasking (ETO for Multitask Optimization in \cite{tan2021evolutionary}), stressing on the algorithmic perspective, b) a manifold taxonomy based on three different pivotal axis: knowledge sharing pattern adopted (implicit or explicit), dynamic nature of the solving schemes (static or adaptive) and the design template of the search algorithm (MFO and MM), c) a critical analysis of the methodological trends followed by researchers when designing and implementing EM-based methods; and d) an insightful discussion around challenges and opportunities fully focused on EM, in which we deal with topics ranging from possible applications, algorithmic enhancements and benchmarking issues.

On the other hand, authors in \cite{xu2021multi} present a brief overview of the work done in the last five years in the field of multitask optimization and EM. Along with the definition and basic concepts of multitask optimization, authors provide a review of the field focusing on different concepts such as encoding schemes, parameterization strategies, knowledge sharing patterns, when and what to transfer, and evaluation and selection strategies. After that, authors explore some additional aspects such as algorithm frameworks, many-tasking optimization problems or similarity measures among tasks. Finally, authors analyze the main applications of EC methods up to now, and they highlight some potential future works in the field. Although it is an interesting survey, it differs substantially from the one presented in this paper. First, our manuscript surveys the EC field using an alternative pivotal criterion, employing our proposed manifold taxonomy based on three axis as the backbone guiding our critical discussion on the field. This taxonomy allows us to examine the work done in an intuitive and homogeneous way, easing the conception of a wide understanding of the current status of the field. We also provide a critical review of theoretical works published so far on EM, which are a key factor to properly understand the evolution of the field over the years. Furthermore, the taxonomy-guided structure of our survey also facilitates the analysis of methodological trends adopted by researchers, providing a priceless contribution to newcomers. Lastly, along with the already mentioned components, significant effort in our paper is devoted to a prospective on the most urgent and interesting challenges and opportunities in the area, which should drive research invested on EM in the upcoming years. This last part is among the principal strengths of our survey, inviting the reader to think beyond conventional paradigms and clearly detaching its discourse from previously published works.

The remainder of this manuscript is structured as follows: Section \ref{sec:essentialconcepts} briefly poses the essential concepts of EM and introduces the reader to the main approaches used so far to face this paradigm. After that, Section \ref{sec:bibl} is dedicated to describe both followed bibliographic method and research questions that have guide our literature analysis. Next, Section \ref{sec:whereweare} delves into the survey itself, departing from the presentation of the taxonomy criteria, to arrive at a careful examination of the recent bibliography related to EM. Equally important is the methodological overview done in Section \ref{sec:metho}, bringing to the fore the main methodologies followed in the different phases of EM algorithmic development. Section \ref{sec:whereweshouldgo} gravitates on the current limitations and discusses several challenges stemming therefrom. Finally, Section \ref{sec:conc} concludes our survey with a summary of the main conclusions and an outline towards the future of this exciting field. 

\section{Evolutionary Multitasking: Definition and Essential Concepts} \label{sec:essentialconcepts}

As introduced before, multitasking is devoted to the simultaneous solving of different optimization problems or tasks. It is important to emphasize at this point that the main goal of this paradigm is to find a promising solution to each of the problems at hand. This specific TO category is featured by an omni-directional knowledge sharing among tasks, potentially reaching a synergistic push between the problems being tackled \cite{gupta2017insights}. In this way, multitask optimization sinks its roots in the premise that these complementarities among tasks lead to a competitive advantage over the case where the same problems are solved in isolation, either in terms of the optimality of the discovered solutions, or in terms of convergence and consumption of computational resources. 

Mathematically, a multitask optimization scenario consists of $K$ optimization tasks $\{T_k\}_{k=1}^K$, which are to be simultaneously solved. In this way, this environment can be characterized by the existence of as many search spaces $\Omega_k$ as tasks. Furthermore, each $k$ task has its own fitness function (objective) $f_k : \Omega_k \rightarrow \mathbb{R}$, where $\Omega_k$ is the search space over which $f_k(\cdot)$ is defined. Assuming that all problems should be maximized, the main objective of multitask optimization is to discover a set of solutions $\{\mathbf{x}_1^{\ast},\dots,\mathbf{x}_{K}^{\ast}\}$ such that $\mathbf{x}_{k}^{\ast} = \arg \max_{\mathbf{x}\in\Omega_k} f_k(\mathbf{x})$.

\begin{figure}[h!]
	\centering
	\includegraphics[width=\textwidth]{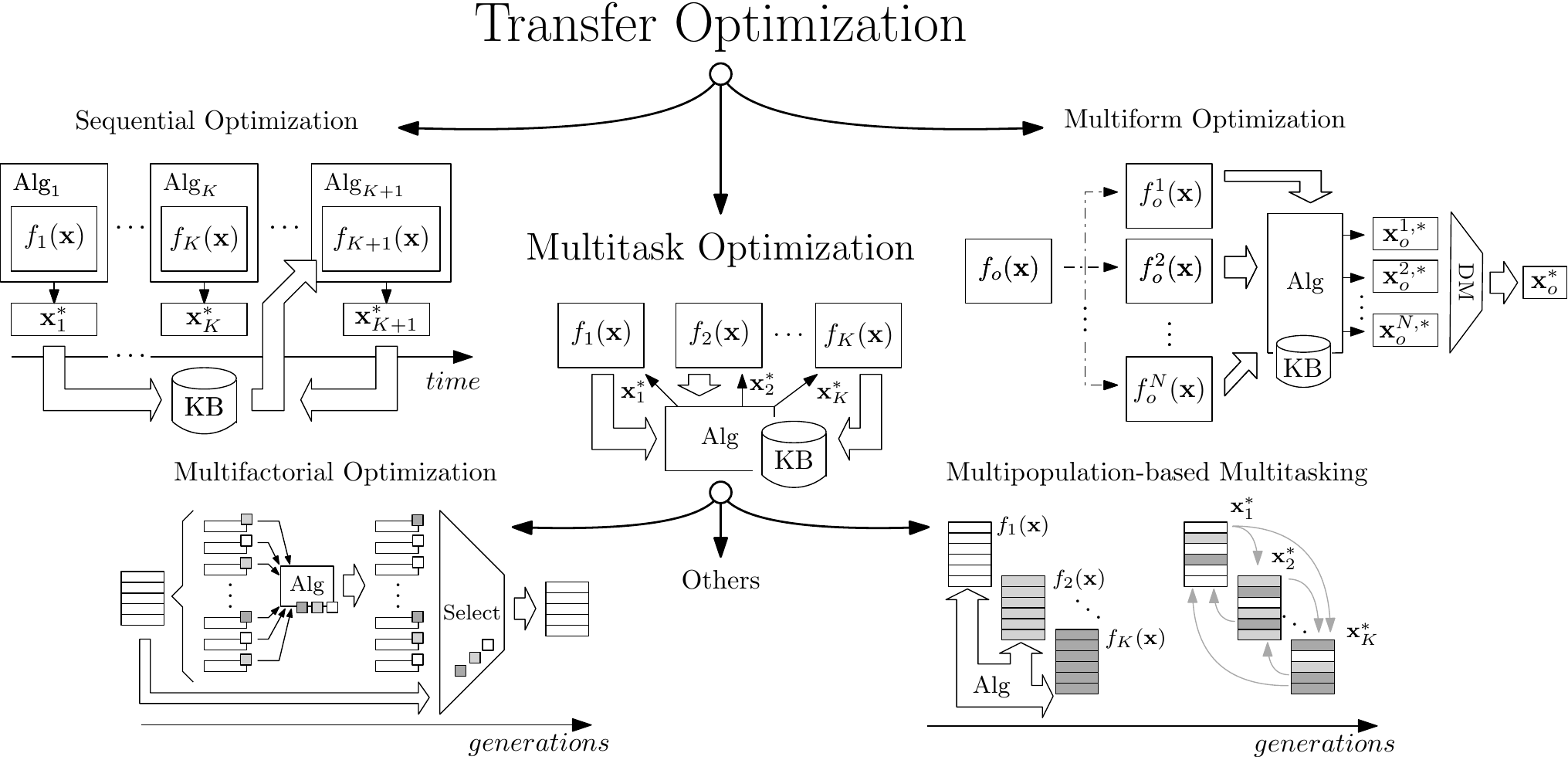}
	\caption{Schematic diagram showing the different ways Transfer Optimization can be realized, along with the two main family of algorithms by which multitask optimization can be approached using concepts from Evolutionary Computation.}
	\label{fig:general}
\end{figure}

We now shift our focus to EM, in which two main characteristics have stimulated researchers to deal with multitask optimization scenarios by means of evolutionary search operators. On the one hand, the intrinsic parallelism that brings a population of individuals which evolve together is well suited to deal with concurrent problems. In fact, several papers have already highlighted the benefits of this structure for dynamically unveiling synergistic relationships between tasks \cite{ong2016evolutionary,bali2017linearized}. On the other hand, the continuous exchange of genetic material along the evolutionary search allows all tasks to benefit from each other \cite{louis2004learning}. Considering the formulation introduced above, there are several ways for dealing with multitasking environments through the prism of EM, being two the most used approaches in the state of the art (schematically depicted in Figure \ref{fig:general}): 
\begin{itemize}[leftmargin=*]
\item The execution of a single search process over a unique population $\mathbf{P}=\{\mathbf{x}^p\}_{p=1}^P$ that contains the solutions to all problems, and that fosters the exchange of information among them through the application of crossover operators (as in e.g. Multifactorial Optimization, MFO). In this case, an aspect of paramount importance is that each solution $\mathbf{x}^p$ in the population should be evolved over an unified search space $\Omega_U$. Thus, each independent search space $\Omega_k$ belonging to task $T_k$ can be translated to $\Omega_U$ by means of an encoding/decoding function $\xi_k: \Omega_k\mapsto \Omega_U$. For this reason, each individual $\mathbf{x}^p\in \Omega_U$ in $\mathbf{P}$ should be decoded to yield a task-specific solution $\mathbf{x}_k^p$ for each of the $K$ tasks. In this context, the appropriate encoding strategy used for the individuals and the capability of the designed unified search space to represent all solutions $\forall \Omega_k$ is crucial for an effective knowledge transfer between tasks. Specifically, the formulation of $\Omega_U$ should be consistent with the level of overlapping among problems being solved. 

\item The deployment of several search processes that run in parallel, one for every task under consideration, which exchange information periodically as per a defined knowledge sharing policy (as in e.g. Multipopulation-based Multitasking, MM). In this case, each search process operates on a task-specific population $\mathbf{P}_k=\{\mathbf{x}_k^p\}_{p=1}^{P_k}$, whose size $P_k$ and search operators can be particular for task $T_k$ and hence, differ from those used for other concurrent tasks. In accordance with previous notation, $\mathbf{x}_{k}^p\in\Omega_k$ $\forall p\in\{1,\ldots,P_k\}$. In this case, the exchange of information is usually made in terms of solutions eventually exchanged between populations belonging to different tasks, so that a mapping function $\Gamma_{k,k'}:\Omega_k\mapsto \Omega_{k'}$ is needed to translate an individual $\mathbf{x}_{k}^p$ to the search space of task $T_{k'}$. This mapping function can be defined and particularized per every task pair or, instead, can rely on an intermediate unified search space, such that $\Gamma_{k,k`}(\mathbf{x}_{k}^p) = \xi_{k'}^{-1}(\xi_k(\mathbf{x}_{k}^p))$, with $\xi_k(\mathbf{x}_{k}^p)\in\Omega_U$.
\end{itemize}

Based on the work published by Ong and Gupta in \cite{ong2016evolutionary}, we can measure the overlap of two problems based on the amount of variables in the task-specific solution space which have the same phenotypical meaning. Thus, three different superposition levels can be identified depending on the amount of overlap in the phenotype space of the optimization tasks: 1) \emph{complete overlap}, when tasks to solve are distinguished only on their task-specific auxiliary variables; 2) \emph{partial overlap}, when problems share some characteristics, or tasks in which the distribution of variables is similar; and 3) \emph{no overlap}, when problems to be tackled do not share any aspect of their structure. In any case, despite the relevance of the level of superposition when designing EM approaches, it is important to be aware that in many real applications it is not possible to measure the level of complementarity among tasks being solved without actually solving them \cite{gupta2017insights}.

Having introduced these concepts, it is appropriate to highlight that there is a common point of agreement in the related community, which states that EM was only materialized by means of MFO until late 2017 \cite{da2017evolutionary}. From that moment on, this incipient branch of TO has gathered a growing amount of contributions centered on the proposal of new EM solvers. Nowadays, it is widely agreed that two are the most recurring approaches for dealing with EM environments: MM and MFO, which conform to the two main design trends described above. 

On one hand, we can define MM approaches in a generalist way as techniques organized by different populations, in which each deme is devoted to the resolution of one specific task. MM can be heterogeneous, giving rise to different solving strategies relying on evolutionary and/or swarm intelligence heuristics or knowledge sharing protocols. Among these strategies, the one known as coevolutionary optimization (CoEV, \cite{paredis1995coevolutionary}) is arguably the most frequently used today, in which knowledge sharing among populations (in terms of e.g., member migration or intra-deme crossovers) helps the evolution of each task. Examples of MM techniques are the multitasking multi-swarm optimization proposed in \cite{song2019multitasking}, the coevolutionary multitasking scheme introduced in \cite{cheng2017coevolutionary} or the coevolutionary variable neighborhood search presented in \cite{osaba2020coevolutionary}.

On the other hand, the design of MFO techniques hinges on the definition of four different albeit interrelated specific concepts for each solution $\mathbf{x}^p\in\Omega_U$ of the single population $\mathbf{P}$ over which the search is performed:
\begin{itemize}[leftmargin=*]
\item \textit{Concept 1 (Factorial Cost)}: the factorial cost $\Psi_k^p\in\mathbb{R}$ of an individual $\mathbf{x}^p\in\mathbf{P}$ is equal to its fitness value $f_k(\mathbf{x}_k^p)$ for a given task $T_k$, which can be computed after decoding $\mathbf{x}^p$ to $\mathbf{x}_{k}^p$ via $\xi_k(\cdot)$. Each member of the population has a list $\{\Psi_1^p,\Psi_2^p,\dots,\Psi_K^p\}$ of factorial costs, each one associated with an optimization task $T_k$.

\item \textit{Concept 2 (Factorial Rank)}: the factorial rank $r_k^p\in\mathbb{N}$ of an individual $\mathbf{x}^p$ for task $T_k$ is the position of this member within the whole population sorted in ascending order of $\Psi_k^p$. Every individual also counts with a factorial rank list $\{r_1^p,r_2^p,\dots,r_K^p\}$.

\item \textit{Concept 3 (Scalar Fitness)}: the scalar fitness $\varphi^p$ of $\mathbf{x}^p$ is computed based on the best factorial rank among the optimization tasks, i.e., \smash{$\varphi^p = 1/\min_{k \in \{1...K\}}r_k^p$}. This value is used for comparing individuals in a MFO algorithm.

\item \textit{Concept 4 (Skill Factor)}: denoted as $\tau^p$, the skill factor is the task index in which member $\mathbf{x}^p$ performs best, that is $\tau^p = \arg \min_{k\in\{1,\ldots,K\}} r_k^p$.
\end{itemize}

The above four concepts are the cornerstone on which all MFO techniques rely. In fact, these definitions are used for different purposes, such as 1) deciding how population individuals interact with each other; 2) determining which solutions survive in the population between successive generations; 3) assigning tasks to individuals; or 4) classifying and sorting the whole population. With all this, these four concepts (either in their seminal form or in modified formulations, such as those proposed in \cite{osaba2020mfcga,osaba2020multifactorial}) have led to several efficient MFO techniques for solving multitasking scenarios. 

Furthermore, it is interesting to highlight here that two different knowledge sharing strategies can be found in EM methods, which can be approached as per the level of explicitness of the exchanged knowledge with respect to the evolved solutions. As such, \textit{implicit transfer} refers to those cases where knowledge sharing is materialized through search operators, such as crossover functions. An example of implicit genetic transfer is the assortative mating used in most MFO techniques. By contrast, \textit{explicit knowledge transfer} is conducted by migrating complete solutions from one task to another, which is often adopted in multipopulation schemes. Furthermore, it also be noted that explicit transfer could also be materialized through the use of mapping functions for transforming solutions before transferring, or by making use of Estimation of Distribution Algorithms (EDA \cite{larranaga2001estimation}). We will revolve around these alternative paths for knowledge transfer when discussing our prospective on the field in Section \ref{sec:whereweshouldgo}.

Notwithstanding the proven efficiency of EM solvers (including those related to MFO), it is appropriate to finish this section by underscoring that multitasking has been the focus of diverse debates questioning the efficiency of techniques proposed to date. Today, it is a clear consensus regarding the paramount relevance of the correlation among tasks to solve. The existence of these interrelationships is essential for positively capitalizing the shared knowledge over the search. Many studies have analyzed from different perspectives the similarities and possible synergies among problems \cite{gupta2016landscape}. However, in many practical environments it is not possible to quantify the existing complementarity among tasks in a preemptive fashion, without any knowledge of the optimal solution to each problem under consideration. This noted fact creates a latent problem for multitasking solvers, as the sharing of genetic material among non-related tasks is known to potentially lead to performance downturns. This phenomenon is known by the community as \textit{negative transfer} \cite{bali2019multifactorial}, and has motivated a significant research upsurge towards alternative EM methods capable of avoiding and/or counteracting its effects in the convergence of the multitasking search. Such alternative methods will be reviewed and discussed in Section \ref{sec:whereweare}. 

For the sake of a solid understanding of the EM paradigm, the following two sections clarify the main differences between multitask optimization, multi-objective optimization (Subsection \ref{sec:MOO}) and multitask learning (Subsection \ref{sec:ML}).

\subsection{Multi-objective Optimization versus Multitask Optimization} \label{sec:MOO}

\begin{figure}[h!]
	\centering
	\includegraphics[width=\textwidth]{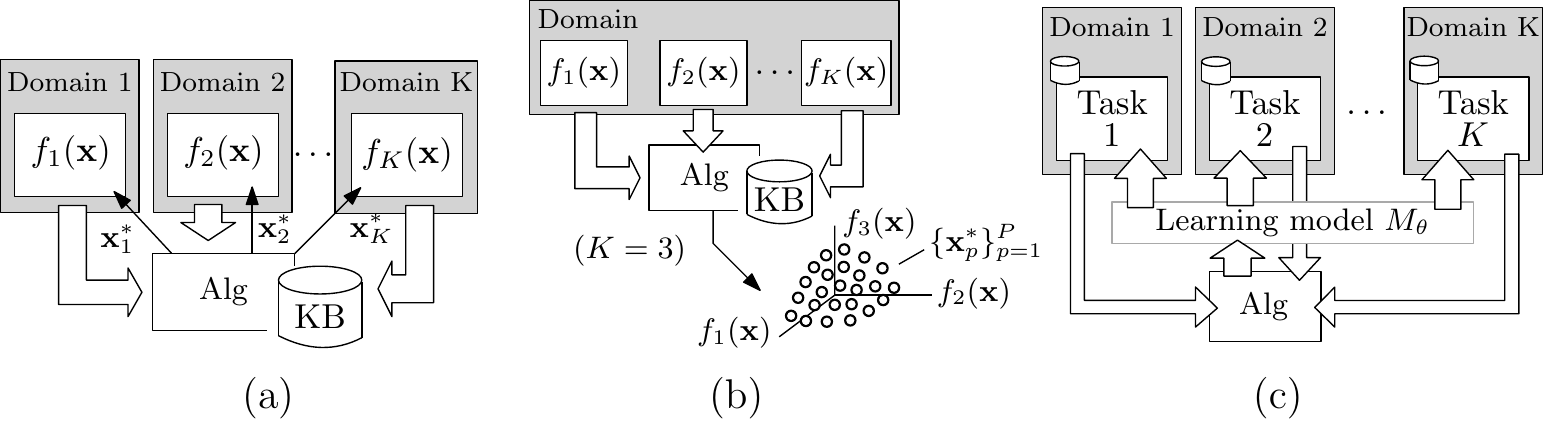}
	\caption{Diagram showcasing the core differences between (a) multitask optimization; (b) multi-objective optimization; and (c) multitask learning.}
	\label{fig:multiobj_multitask}
\end{figure}

An insightful reader can immediately relate EM to Multi-objective Optimization (MOO) paradigm which, when approached via evolutionary computation, span the wide family of multi-objective evolutionary algorithms. Indeed, it is possible to discern a conceptual overlap between both EM and MOO, since both aim at the optimization of a set of objective functions. However, as shown in Figures \ref{fig:multiobj_multitask}.a and \ref{fig:multiobj_multitask}.b, these paradigms are completely separated from each other. On the one hand, EM aims to leverage the inherent parallelism enabled by a population of individuals for exploiting the synergies among related or unrelated tasks defined in different domains, each with its own solution space $\Omega_k$ that potentially requires an encoding/decoding function for knowledge transfer. Moreover, EM also pursues the discovery of the best solution for every task. On the contrary, the goal of MOO is to find a set of solutions that differently balances between several conflicting objectives, defined over a single domain (and hence, over a single search space). In other words, MOO assumes the existence of a Pareto trade-off between the objectives, for which the devised MOO algorithm produces an estimation in the form of a set of possible solutions. Therefore, there is no \emph{unique} solution to each problem, but rather different solutions that meet every objective to a certain degree. In fact, EM setups where the tasks themselves are MOO problems can be found in the literature \cite{gupta2016multiobjective}.

\subsection{Multitask Learning versus Multitask Optimization}
\label{sec:ML}

Multitask learning and multitask optimization work on similar scenarios, in which a set of solutions $\{\mathbf{x}_{1}^\ast,\dots,\mathbf{x}_{K}^\ast\}$ is sought for a set of tasks $\{T_k\}_{k=1}^K$. However, they mainly differ in terms of the optimization target, and the way in which the knowledge transfer is carried out. In multitask learning, the goal is to yield a model $M_{\theta}$ (with $\theta$ representing the parameters of the model) such that it can tackle the goal imposed by different tasks (e.g. classification of images of diverse kind). Here, the challenge is to determine a model structure and a value of their constituent parameters that best favors not only a good performance on every task under consideration, but also the exploitation of the synergies between modeling tasks. This dual functionality sought in multitask learning underlies beneath the design of multi-headed neural networks with shared losses trained via backpropagation, which are arguably the most utilized approach in the field: on one hand, sharing part of the neural architecture permits that part of the knowledge is common to all tasks, whereas the definition of a shared loss function ensures that the optimization of the parameters of the network is driven by the performance over all tasks.

This being said, there is a clear connection between multitask learning and multitask optimization, in the sense that multitask learning can be stated as a multitask optimization problem, provided that 1) solutions $\{\mathbf{x}_k^\ast\}_{k=1}^K$ elicited by multitask optimization represent the parameters of a model, and 2) solutions are constrained to part of their genotype being shared among tasks, so that they jointly embed a single model. This last constraint can be overridden so as to produce a set of models that collaborate together to solve several learning tasks more efficiently than in isolation. In this case, each optimization task would aim to seek the parameters of the model that best performs over the defined modeling problem for the task, and implicit/explicit knowledge transfer mechanisms used in EM could be effectively employed in place to transfer the knowledge learned in a certain task to another. All in all, multitask optimization must be conceived as a possible way of approaching multitask learning, but not the only one whatsoever.

\section{Bibligraphic Method and Guiding Research Questions}
\label{sec:bibl}

With the main intention of conducting a thorough and valuable survey, we dedicate this section to describing the research methodology followed for conducting this review paper. For properly gathering all the scientific material published around this incipient research field, several searches have been iteratively conducted using the most reputed and well-known scientific databases: the Clarivate Analytics Web of Science, Scopus and Google Scholar. At this moment, it is interesting to mention that due to the fact that Evolutionary Multitasking is a field that has not reached maturity, multiple terms have been employed for a systematic discovery of the published papers. The main reason for this situation is the non-existence of a general and well-accepted terminology for certain related concepts. For this reason, the next terms have been used for digging up all the produced material: "Transfer Optimization", "Multitasking", "Evolutionary Multitasking", "Evolutionary Multitask Optimization", "Multifactorial Optimization", "Multifactorial Evolutionary Algorithm" and "Multipopulation-based Multitasking". 

Furthermore, after carrying out a first sweep using these terms, an exhaustive analysis has been performed paper by paper, in order to identify its adequacy for the present study. Once again, this situation is also a direct consequence of the youth of this field, which is why there are articles that using inaccurate terminology can lead to ambiguities. Having applied this second filter, and after the final selection of the articles to review, each manuscript has been categorized using the following criteria: knowledge transfer strategy employed, capacity of the solving approach for analyzing the negative knowledge and principal algorithmic scheme. Finally, and going deeper into the bibliographic analysis carried out, the three main research principles that have guided our investigation have been the following ones:

\begin{itemize}
	\item To determine which are the predominant methodological patters that guide the current algorithmic developments, in terms of knowledge sharing among solving tasks, and adaptation to negative transfer phenomenon. 
	\item To clearly identify which are the principal used mechanisms and operators for the evolution of Evolutionary Multitask Optimization Methods, both in MFEA and MM related schemes.
	\item To establish a strong basis for a prescription of methodological improvement areas and opportunities for future research, which we conduct in Section \ref{sec:metho} and Section \ref{sec:whereweshouldgo}.
\end{itemize}

Embracing the method described in these paragraphs, the next section is devoted to presenting the taxonomy elaborated on the topic at hand, EM, and to outlining the main progresses conducted by researchers and practitioners to date.

\section{Taxonomy and Literature Review on Evolutionary Multitasking}
\label{sec:whereweare}

As mentioned in the introduction of this paper, the research activity produced around EM is growing at a remarkable path since the first formulation of this vibrant paradigm. In any case, all the work done specifically around EM has not been organized yet in a scientific paper, arising the need for properly conducting a work of this characteristics. Indeed, this is precisely the main objective of this section, in which we systematically review the most important works published to the date in the field of EM. 

\begin{figure}[h!]
	\centering
	\includegraphics[width=0.85\columnwidth]{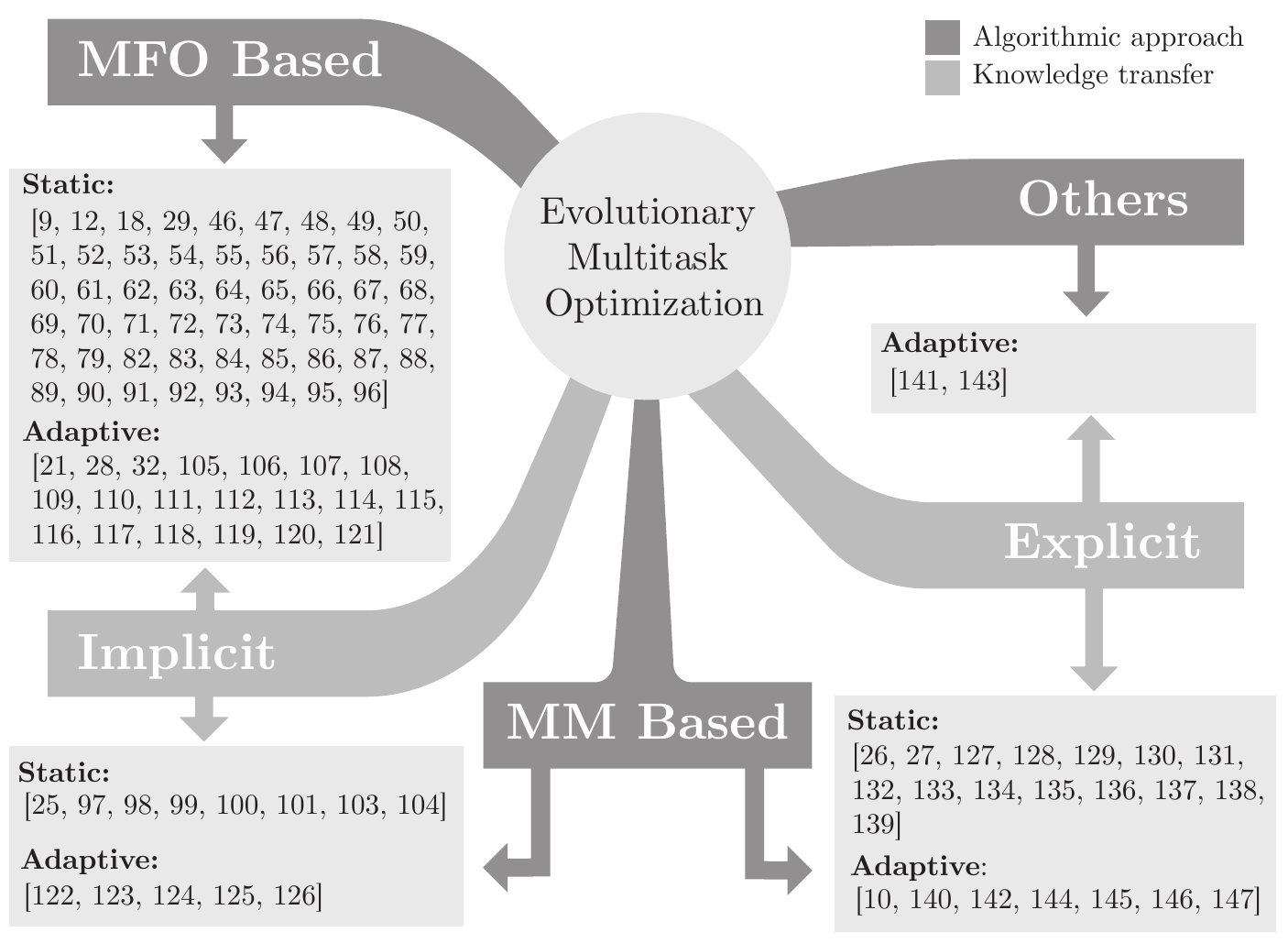}
	\caption{Taxonomy of the literature related to Evolutionary Multitask Optimization reviewed in this survey. A two-level classification has been made depending on the knowledge transfer scheme used and the adaptability of the proposed methods. Furthermore, solvers have been categorized into MFO and MM based ones, with an additional 'other' classification.}
	\label{fig:tax}
\end{figure}

In order to appropriately guide this section, we first present in Figure \ref{fig:tax} a taxonomy covering all the studies contemplated in this review section. For organizing this taxonomy, and subsequently this section, we have classified all the published material using a two-level approach. First, we have deemed the knowledge transfer strategy employed by the proposed solving method (implicit or explicit). The second level regards to the capacity of the method to proactively analyze the negative knowledge sharing among tasks and dynamically react to this issue, seeking to reduce its impact in the algorithmic search. On the one hand, if the solving approach does not include any analyzing mechanism, we have considered it a \textit{static}. On the other hand, we consider an algorithm as \textit{adaptive} if it not only employs this kind of analyzing strategies but adapt its structure to the unveiled synergies among tasks (by modifying the parameters of the algorithm, for example). Lastly, once this categorization is conducted, we have further classified the published contributions taking as reference the algorithmic approaches used and proposed by researchers and practitioners. In this regard, we have considered \textit{MFO based schemes}, \textit{MM based approaches} and \textit{other methods}.

With all this, this taxonomy sorts the literature according to these algorithmic schemes, being also valuable for distinguishing at a short glimpse those areas in which the community has so far place most of their attention. This literature overview, together with the methodological review conducted in Section \ref{sec:metho}, settle a stepping stone towards the critical discussion that will be held in Section \ref{sec:whereweshouldgo} around the main limitations, opportunities and challenges that bring this area.

\subsection{Theoretical Studies on Multitask Optimization}
\label{sec:theo}

As mentioned, this whole section will revolve around the systematic overview of all the work done up to now on Evolutionary Multitask Optimization. This overview has been conducted though the perspective of both algorithm proposals and their knowledge sharing patterns. Nevertheless, we would make a big mistake if we leave aside the large number of paramount articles which have contributed in a crucial way to the establishing, advancement and understanding of this field. Specifically, we refer to the theoretical papers, which address the knowledge area from a less applied point of view, in order to understand in an adequate way the ins and outs of the field. These works are essential for establishing in the community the main pillars that make the research stream can advance in an orchestrated and efficient way.

Probably, the most valuable paper in this context is this published by Ong et al.~in \cite{ong2015towards}, which is devoted to the introduction and presentation of EM field. This works is a cornerstone in the research community, establishing the basic concepts that have guided all the work conducted in last years. Apart from this influential and pioneering contribution, several remarkable theoretical works have been published on EM delving of different aspects such as the influence of complementarities between function landscapes on the search performance \cite{gupta2016measuring,gupta2016landscape}, or just highlighting the main ingredients that make this knowledge stream interesting for the research community \cite{gupta2016genetic}. Further works on EM from a theoretical point of view can be found in \cite{gupta2017emerging,ong2016evolutionary}.

It is interesting to mention again at this point the study published by Gupta et al in \cite{gupta2015multifactorial}. That paper is not only significant for introducing to the community the most important method to date, MFEA, but also for establishing the principal wickers that make up MFO. As will be demonstrated in Section \ref{sec:ImplicitStatic} and Section \ref{sec:ImplicitAdaptive}, both MFO and MFEA have been the source of inspiration for an abundant number of valuable works. As part of the work carried out, there are several published papers that have also delved into theoretical aspects of these paradigms. In the recent \cite{huang2020analysis}, for example, an analysis on the efficiency of MFEA is carried out. Main objectives of that study are twofold: to theoretically unveil why MFEA based methods perform better that classical techniques, and to provide some findings on the parameter setup of MFEA algorithm. In the recent \cite{xu2020parameter}, the impact of three different MFEA parameters is analyzed: probability of individual learning, probability of intra-crossover and probability of inter-crossover. In \cite{wang2019rigorous}, a rigorous analysis is carried out on the relationship of MFEA and the conceptually similar multipopulation evolution models. To do that, authors make an in-depth comparison on their performance and working procedures. A similar study is also proposed in \cite{hashimoto2018analysis}, revolving around the idea of the relationship among MFEA and island-based models. Interesting is also the brief study proposed in \cite{zhou2018study}, focused on presenting insights in the measure of task relationship in MFEA. In \cite{yao2021novel}, the efficiency of the binary tournament selection criteria used in the Multiobjective variant of the MFEA is studied, proposing additional selection strategies. Further examples of theoretical studies can be found in \cite{wang2020analysis}, focused on analyzing the influence of the order of solution variables in multitasking environments; or in \cite{bai2021multitask}, devoted to the analysis of convergence in evolutionary multitasking scenarios compared to the conventional single task optimization.

Significant is also the position paper published by Gupta and Ong in \cite{gupta2019back}. The main objective of that research is to return to the roots of Evolutionary Computation. From here, authors provide an interesting review of the field for properly understand the inspirations of what can be classified under the umbrella of \textit{multi-X evolutionary computation} concept. Thus, multitasking is again analyzed in this paper from its theoretical perspective. 

Finally, it is interesting to mention in this category the studies described in both \cite{yuan2017evolutionary} and \cite{da2017evolutionary}. These reports contribute to the EM field by introducing some valuable test problems for both single-objective MFO and multi-objective MFO. Main intention of the authors of that works it to present to the community some heterogeneous benchmarks and baseline results, in order to use them for subsequent studies.

\subsection{Implicit Knowledge Transfer Based Static Solvers}
\label{sec:ImplicitStatic}


As mentioned in the previous Section \ref{sec:essentialconcepts}, implicit transfer is often materialized through the application of dedicated search operators such as crossover functions. The principal standard-bearer for this type of transfer is known as \textit{assortative mating} which is used in most of MFO techniques. The first paper revolving around \textit{assortative mating} procedure is the same work which also introduces Multifactorial Optimization paradigm \cite{gupta2015multifactorial}. This paper became instantly in a reference paper, not only because of the introduction of MFO concept, but also for the formulation of the most used and influential EM technique: the Multifactorial Evolutionary Algorithm (MFEA). From that moment on, many diverse adaptations and applications of the canonical MFEA has been proposed in the literature. In \cite{yuan2016evolutionary}, for example, first discrete adaptation of MFEA is proposed, using as benchmarking problems four well-known permutation-based combinatorial optimization problems: Traveling Salesman Problem, Quadratic Assignment Problem, Job-Shop Scheduling Problem and Linear Ordering Problem. After that pioneering study, multiple additional discrete adaptations of the method have been proposed, such as the one focused on solving the Capacitated Vehicle Routing Problem in \cite{zhou2016evolutionary} or the series of works published principally by Thanh and Binh for the facing of clustered shortest path tree problems \cite{thanh2020efficient,binh2018effective,thanh2018effective,thanh2020two,dinh2020multifactorial,hanh2021evolutionary,binh2021bi}.

Several adaptations of MFEA have been also proposed for efficiently deal with real-world optimization problems, such as the permutation-based MFEA proposed in \cite{bao2018evolutionary} with cloud computing service composition purposes. A quite related approach was presented in \cite{wang2019evolutionary}, devoted in that case to the efficient semantic web service composition. In \cite{liang2020multi}, authors develop a MFEA embedded with a greedy-based allocation operator for solving large-scale virtual machine placement problem in heterogeneous environment. An additional interesting application of MFEA has been recently proposed in \cite{martinez2020simultaneously}, with the main goal of simultaneously evolving concurrent deep reinforcement learning models.

Due to the success of MFEA, multiple variants of this method have been recently proposed. These methods also rely on implicit transfer strategies for the genetic material sharing, using the majority of them the above pointed \textit{assortative mating}. In \cite{ding2017generalized}, for example, authors introduced the named as Generalized MFEA. The main reason for the formulation of this technique is that MFEA experiences performance downturns when dealing with tasks with different dimensions, or problems whose optima do not lie in the same region of the solution space. The Generalized MFEA try to overcome these issues by implementing two different mechanisms related to decision variable translation and shuffling.

Another interesting variant of MFEA is developed in \cite{yin2019multifactorial}. The improved MFEA proposed in this work explores the integration of a novel cross-task implicit transfer operator, which is based on a search direction instead of an individual. The main objective of this method is to accelerate the convergence of the search process, especially in environments where the optima of tasks are far from each other. In another vein, authors of \cite{huynh2020multifactorial} modeled an interesting hybrid MFEA which combines both MFEA and the Linkage Tree Genetic Algorithm. In \cite{da2016boon}, a variant of MFEA coined as \textit{polygenic evolutionary algorithm} is designed, which curtails the cultural issues of the evolutionary procedure in the models of multifactorial inheritance. The main objective of that work is to understand the importance of both \textit{assortative mating} and \textit{vertical cultural transmission} towards effective evolutionary multitasking. Further interesting MFEA variants have been proposed in \cite{lian2019improve} by means of the coined as (4+2) MFEA; in \cite{zhou2020mfea} introducing the MFEA with Individual Gradient mechanism for enhancing knowledge transfer; and \cite{binh2020multifactorial} with MFEA with Priority-based Encoding. Further applications and variant of the MFEA can be found in \cite{wang2020multifactorial} and \cite{xue2020affine}.

Soon after the proposal of the single-objective, same authors that introduce MFEA proposed the Multi-Objective MFEA (MO-MFEA, \cite{gupta2016multiobjective}), which automatically entailed an inflection point in the related research community. It should be noted here that this Multi-Objective MFEA also employs the \textit{assortative mating} procedure for implicit knowledge sharing purposes. Furthermore, this specific method has been already used in a heterogeneous range of applications, such as for dealing with the multi-objective pollution-routing problem \cite{rauniyar2019multi} or for the electric power dispatch \cite{liu2020multitasking}. A further application of MO-MFEA was presented in \cite{yang2018multitasking} for solving operational indices optimization.

Furthermore, improved variants of the basic method have been already proposed, such as the one modeled in \cite{yang2017two}. In that paper, authors introduce a MO-MFEA with a two-stage \textit{assortative mating} method. This procedure introduced a preliminar division of the decision variables into diversity-related variables and convergence-related variables. After this first step, both types of variables undergo the \textit{assortative mating}. Another adaptation was developed in \cite{mo2017multi}, coined as decomposition-based MO-MFEA (MFEA/D-M2M). The main ingredient that characterizes this method is the adoption of a M2M approach for decomposing multi-objective optimization problems into multiple constrained sub-problems. The main goal of this procedure is to enhance the diversity of population and convergence of sub-regions. Also valuable is the study carried out in \cite{zhou2020multi}, focused on the resolution of the well-known multi-objective vehicle routing problem with time windows using an improved MO-MFEA by integrating bone route and large neighborhood local search. Furthermore, authors of \cite{tuan2018guided} introduced a so-called Guided Differential Evolutionary (DE) MO-MFEA. Two are the main novel ingredients of this method: a) an improved crossover operator using guided differential evolution, and b) a modified Powell mechanism for mutation operations. An additional improved version of the MO-MFEA can be found in \cite{yi2020multifactorial}, devoted to the solving of interval multiobjective optimization problems (cases in which the coefficients in their objectives or/and constraint(s) are intervals). 

It is interesting to highlight the single and multi-objective optimization multifactorial evolutionary algorithm (S\&M-MFEA) proposed in \cite{da2016evolutionary}. The main purpose of that solving scheme is to combine in a single multitasking environment the original single-objective MFEA formulation together with its associated multi-objective reformulation. Finally, in \cite{ma2021enhanced} authors proposed a MFEA with the incorporation of a prior-knowledge-based multiobjectivization via decomposition, with the main goal of building strongly related meme helper-tasks.


Despite the huge success and the contrasted efficiency of MFEA, researchers have rapidly detected the main limitations inherent to the canonical scheme of this method. As reported in previously published papers \cite{bali2019multifactorial}, the main limitation of MFEA is its difficulty for facing potential incompatibilities between different non-related tasks. For dealing with this issue, two principal research streams have been followed by the community up to now. The first one is the development of adaptive methods (as will be seen in Section \ref{sec:ImplicitAdaptive} and \ref{sec:ExplicitStatic}). The other approach is the design of alternative solving schemes. Within this last category, different techniques can be found in the literature that address EM throughout the lenses of MFO but using a different scheme than MFEA. Another limitation of MFEA is that it resorts to non-structured populations, even though we assume that such a structure exists. Thus, a more advanced and sophisticated structures could favor the design of alternative search operators, promoting a more controlled exchange of knowledge between related and non-related tasks.

Therefore, to overcome these limitations, practitioners have taken a step forward, proposing novel mechanisms which have led to the proposal of numerous methods, based on the essential concepts of the MFO. The first alternative MFEA scheme was proposed in \cite{gupta2015evolutionary}, just some months later that the seminal work presenting the canonical MFEA. The main motivation that led the conduction of that work is to demonstrate that the practicality of population-based bi-level optimization could be enhanced by deeming the paradigm of EM within the search process. To do that, authors embedded the principal MFO concepts into the scheme of the well-known Nested Bi-Level Evolutionary Algorithm, giving rise to the coined as N-BLEA. Some months later, Sagarna and Yew-Soon introduced in \cite{sagarna2016concurrently} a MFO method for search-based software test data generation. In an attempt of leveraging the knowledge from different sources and enhance the search process, authors of that work proposed a MFO algorithm which bases the complete search procedure in mutation operations. Thus, authors evince that the selection operator and the preference relation used to compare individuals allow to inter-task knowledge transfer for an effective search.

Also proposed shortly after the introduction of MFEA, we can find in \cite{feng2017empirical} an interesting work delving in the main concepts of MFO. The principal goal of that work is to explore the generality of the MFO paradigm, employing different population-based schemes. To do that, authors proposed the first multifactorial formulations of the hugely famous particle swarm optimization (PSO, \cite{kennedy1995particle}) and differential evolution (DE, \cite{price2006differential}). Regarding the knowledge sharing strategies used in these DE- and PSO-based methods, they also employ the widely used concept of \textit{assortative mating}, adapted to the mechanisms of the metaheuristics at hand. Thus, they also rely on implicit transfer mechanisms. Indeed, that interesting work has served as guiding light for subsequent studies, such as the one conducted in \cite{zhou2019towards}, in which the performance of different mutation strategies in the knowledge transfer of multifactorial DE is studied. Furthermore, this same method is used as base in the remarkable investigation carried out in \cite{tang2020regularized}, which main goal is to identify the essential characteristics of tasks landscapes through the implementation of an inter-task evolutionary mechanism in the low-dimension subspace. Another example is the work proposed in \cite{chen2020evolutionary}, in which a MFO based PSO is proposed for feature selection purposes.

Another example of this scientific trend is the Multifactorial Cellular Genetic Algorithm (MFCGA, \cite{osaba2020multifactorial,osaba2020transferability}), which hybridizes the main concepts of MFO with the structural design and behavior patterns of well-known Cellular Genetic Algorithms. Main inspiration of that method is to have a more controlled implicit mating process among different tasks, favoring in this way the exploration and quantitative examination of synergies among the problems being solved. Also interesting is the approach introduced in \cite{xiao2019multifactorial} proposing a multifactorial particle swarm optimization - firefly algorithm hybrid technique. Main feature of this method is that individuals of the population can behave as a particle or a firefly, depending on the search performance. In any case, despite each member of the population can eventually move following each pattern, each individual maintains its nature along the complete execution. Further alternative MFO schemes can be found in \cite{feng2020large}, presenting a method for solving large-scale optimization problems called as evolutionary multitasking assisted random embedding; in \cite{liang2019hybrid}, which introduces a MFO method hybridizing genetic transform and hyper-rectangle search strategies; in \cite{hao2020unified}, which proposed an unified framework of evolutionary multitasking graph-based hyper-heuristic based on MFO concepts; and in \cite{wang2021learning}, which presents a random inactivation based batch many-task evolutionary algorithm, coined as IBMTEA-FCM. Additional MFO inspired techniques can be found in \cite{li2018multipopulation,li2020multifactorial,zhong2018multifactorial,li2021covariance}.

It is worthy to mention that alternative MFO solving schemes have also been proposed for tackling multi-objective optimization problems. In this regard, we can highlight the recent research conducted in \cite{shen2020evolutionary} by Shen et al.~, introducing a novel multitasking multiobjective memetic algorithm for learning Fuzzy cognitive maps, inspired by the principal concepts of MFO. Furthermore, in \cite{li2018evolutionary} a novel multitasking method is proposed, which is fully devoted to the resolution of the sparse reconstruction problem. The method developed on that work was coined as multitasking sparse reconstruction (MTSR), and also relies of MFO concepts such as skill factor, factorial rank and scalar fitness for the multi-objective solving of the problem. It is also interesting to mention the genetic transfer scheme developed for that MTSR method, which is an enhanced variant of the \textit{assortative mating} procedure coined as \textit{Within-Task and Between-Task Genetic Transfer}.

Finally, it is convenient to end this section dedicated to implicit knowledge transfer-based solvers by highlighting a few EM alternatives recently proposed which do not embrace the MFO paradigm. On the contrary, they adopt previously described MM schemes. A representative approach is presented in \cite{song2019multitasking}. In that work, authors develop a dynamic multi-swarm method for EM. In that algorithm, the complete population is divided into as much swarms as task to solve. Furthermore, each subpopulation is divided into different sub-swarm. Thus, within each task subpopulation, a dynamic multi-swarm method is conducted. Furthermore, the knowledge sharing is realized through probabilistic crossover procedures with particles from other tasks groups, giving way to the coevolutionary factor of the method. Moreover, a parallel DE is proposed in \cite{jin2019study}, which introduces knowledge transfer patterns based on the archives of each DE solver. Interesting is also the MM technique developed in \cite{xu2020fireworks}, focused on the multitasking adaptation of the well-known Fireworks Algorithm \cite{tan2010fireworks}; or the method based on genetic programming introduced in \cite{zhang2020preliminary}, in which tasks are solved in dedicated static subpopulations, allowing the crossing among individuals of different demes. Lastly, worth-mentioning is the multi-objective MM method proposed in \cite{wang2021multiobjective}, which adapts the well-known multiobjective optimization evolutionary algorithm based on decomposition (MOEA/D, \cite{zhang2007moea}). In that algorithm, the implicit exchange of genetic material is produced through crossover procedures among individuals specialized on tackling different tasks. To do that, external neighborhoods are generated for each solving problems. Further methods of this category can be found in \cite{xu2020effective} and \cite{liang2020evolutionary2}.

\subsection{Implicit Knowledge Transfer Based Adaptive Solvers}
\label{sec:ImplicitAdaptive}

As mentioned in Section \ref{sec:essentialconcepts} of this work, a significant effort has been conducted by the community for overcoming the problems related to the so-called \textit{negative transfer}. Examples of these alternative schemes are the adaptive EM methods. These instruments are mainly conceived for dynamically calculate the synergies among tasks, and subsequently measure how much knowledge should be transferred across different tasks. Thus, in this section we outline those MFO methods proposed up to now to dynamically cope with the curse of \textit{negative transfers}.


To start with, it is appropriate to mention the recently proposed MFEA-II \cite{bali2019multifactorial}, conceived as the evolved version of the standard-bearer method of the field: MFEA. Thus, two are the main ingredients embedded in the basic MFEA for evolving it to its adaptive variant MFEA-II. First, the parameter which dictates the extent of transfers (RMP) is now codified as a matrix, with a dedicated value for each pair of tasks. Second, this matrix is continuously adapted based on the performance of the multitasking search. It is also noteworthy that this method has been already adapted to discrete problems as can be seen in the recent work \cite{osaba2020dmfea}. Furthermore, same authors that developed the single-objective MFEA-II introduced also its multi-objective version in \cite{bali2020cognizant}. As in the case of the static MFEA, these adaptive schemes also base their knowledge sharing on implicit procedures based on genetic crossover functions.

Another adaptive MFEA is proposed in \cite{zheng2019self}. In that paper, the method is endowed with a self-regulation mechanism. The main objective of this mechanism is to automatically capture the useful knowledge in common of the tasks at hand. For materializing this goal, this approach introduces the concept of ability vector, which substitutes the skill factor $\tau^p$, and which reflects the solutions capability for tackling each of the optimizing tasks. Furthermore, similar authors that proposed MFEA-II in 2019, introduced two years before a Linearized Domain Adaptation MFEA (LDA-MFEA) \cite{bali2017linearized}. This variant can be considered as an adaptive one, since it employs the linear transformation strategy for mapping the landscapes of a simpler tasks to the search space of complex ones. In that way, authors try to conduct efficient knowledge transfer between the problems while being optimized in concert. In the same year 2017, authors in \cite{wen2017parting} proposed a MFEA with parting ways detection and resource reallocation mechanisms. The first of this functionalities is in charge of detecting the occurrence of parting ways at which the sharing of knowledge is being unproductive, while the second mechanism reallocate fitness function evaluation on different types of generated solutions by ceasing the knowledge transfer when parting ways.

Interesting is also the work carried out in \cite{tang2018group}, in which a Group-Based MFEA (GMFEA) is modeled and implemented. The GMFEA has the main characteristic of dividing tasks into different conceptual groups depending on their proved synergy. Thus, GMFEA controls the implicit genetic transfer between problems belonging to same group. The most important feature is that the grouping is performed dynamically, without the requirement of any prior knowledge. Also remarkable is the research recently conducted in \cite{zhou2020toward}. In that paper, authors first explore how diverse kind of crossovers impact on the implicit knowledge transfer in MFEA for solving continuous optimization problems. After that, they introduce a novel MFEA with adaptive knowledge transfer (MFEA-AKT), in which the mating function used for the genetic material sharing is autonomously adapted employing the information gathered on the complete search process. Furthermore, authors in \cite{yao2021self} proposed a simple Self-Adaptive MFEA, which regulates the $rmp$ parameter using a novel inter-task similarity measurement mechanism. Authors used their Self-Regulated MFEA for solving reservoir production optimization problems.

Furthermore, in addition to the above-mentioned multi-objective MFEA-II, a further adaptive variant of the MO-MFEA has been proposed in \cite{binh2019multi}. The specific method implemented in that work is characterized for introducing two novel ingredients: a) the deeming of a set of reference points to determine the diversity of current population (instead of using the crowding distance), and the online adaptation of the Random Mating Probability (RMP) with the intention of improving the genetic transfer of high-similar tasks. A further interesting method of this kind is developed in \cite{yao2020multiobjective}. Specifically, this work is devoted to the implementation of a so-called MO-MFEA with decomposition and dynamic resource allocation strategy (MFEA/D-DRA). A further adaptive version of the MO-MFEA is proposed in \cite{wu2021multitasking}, devoted in that case for the optimal operation of integrated energy systems.


Analogous to static MFO algorithms, researchers and practitioner have also proposed several adaptive multifactorial methods inspired by the main concepts of this EM paradigm. As mentioned before, MFO methods are the main exponents of implicit knowledge transfer-based approaches. We can highlight first the adaptive multifactorial memetic algorithm proposed in \cite{chen2017adaptive}, which congregates a) the use of local search mechanisms influenced by the knowledge learning among problems, b) a re-initialization procedure for overcoming premature convergence issues and c) a self-adaptive parent selection strategy based on search performance. Also valuable is the work conducted in \cite{osaba2020mfcga}, which is focused on developing an adaptive variant of the above mentioned MFCGA. The coined as Adaptive Transfer-guided MFCGA introduces two dynamic ingredients: a) a dynamic reorganization of cellular grids based on search performance and b) a self-adaptive multi-mutation mechanism. A further MFO adaptive variant can be found in \cite{tang2019adaptive}, devoted to the presentation of a multifactorial PSO method with a self-adaption strategy for adjusting the inter-task learning probability. Similar authors proposed in \cite{tang2021multi} an additional adaptive method also based in the well-known PSO, following in this case the MM philosophy and implicit knowledge transfer pattern. Furthermore, in the recent \cite{martinez2021adaptive}, authors proposed an adaptive variant of the MFEA coined as A-MFEA-RL for simultaneously evolving multitasking reinforcement learning scenarios.

As multi-objective alternatives, we can highlight the adaptive multiobjective and multifactorial DE algorithm (AdaMOMFDE) proposed in \cite{wang2019multiobjective}, based on multiple mutation operators which are selected following and adaptive strategy according to their search results. Also significant is the multiobjective and multifactorial subspace alignment and self-adaptive DE (MOMFEA-SADE) recently introduced in \cite{liang2020evolutionary}. Principal ingredients of that method are a) a mapping matrix get by subspace learning and employed for modifying the search space and minimize the impact of negative transfers, and b) a self-adaptive trial vector used on the DE, for generating new solutions influenced by previous experiences. Finally, the work conducted in \cite{zhao2020endmember} revolves around the multitasking adaptive formulation of the MOEA/D. 

Finally, it should be highlighted that, also in this category, several MM solvers have been proposed in recent years in addition to those based on MFO. In this line, it is worth to describe first the coevolutionary multitasking framework proposed in \cite{liaw2017evolutionary}, coined as evolution of biocoenosis through symbiosis (EBS). Inspired by the symbiosis in biocoenosis, EBS is comprised by multiple populations, running in each of them an independent Evolutionary Algorithm. Furthermore, the information exchange among tasks constitutes the so-called symbiosis, and it is conducted through an implicit transfer procedure coined as \textit{Information Exchange through Concatenate Offspring}. Finally, this method introduces adaptive mechanisms for controlling information exchange, mainly based on the search performance. Further works on this method can be found in \cite{liaw2019evolutionary} and \cite{liaw2020evolution}. In \cite{bi2020learning}, an adaptive solver based on genetic programming is proposed, devoted to the dealing of image feature learing. More specifically, this method generates different subpopulations in a dynamic way and the knowledge sharing happens through crossover procedures among individuals of different tasks. 

Additional valuable coevolutive framework is proposed in \cite{chen2019adaptive}, named as many-task evolutionary algorithm (MaTEA). This framework is similar to EBS in terms that it is also featured by having multiple populations governed by an Evolutionary Algorithm, each one dedicated to the optimizing of one tasks. Main characteristic of this MaTEA is an adaptive selection mechanism for choosing suitable assisted task for a given problem based on the accumulated rewards of positive knowledge sharing during the search. Moreover, a genetic material transfer schema via crossover is used for sharing information between problems for improving the efficiency of the search, giving rise to the coevolutionary nature of the method. 

\subsection{Explicit Knowledge Transfer Based Static Solvers}
\label{sec:ExplicitStatic}

All the works mentioned in this paper up to now clearly attest the importance that EM field has in the current scientific community. Furthermore, the intense activity highlighted in previous Sections \ref{sec:ImplicitStatic} and \ref{sec:ImplicitAdaptive} also unveils the importance of MFO in this specific branch of kna travesowledge. In any case, this success cannot overshadow the fact that researchers and practitioners have proposed alternative schemes to MFO to deal with EM environments. Most of these schemes are MM based approaches, principally characterized by embracing explicit knowledge sharing strategies. In this section, we outline the main work conducted in last years around explicit transfer based static solvers. As introduced in previous Section \ref{sec:essentialconcepts}, this kind of knowledge transmission is usually conducted by migrating complete solutions among populations, namely from one task to another one. Additionally, explicit transfer could also be capitalized through the use of mapping functions or making use of EDA-style probabilistic models instead of raw solutions.


Arguably, the most successful alternative trend to MFO paradigm is the one related to MM approaches. Going deeper, most used MM methods fall inside the category known as coevolutionary. These multitasking methods are featured by being composed by multiple populations of individuals, which are usually independently dedicated to the optimization of a single tasks. Thus, the autonomous evolution of these subpopulations together with the punctual sharing of genetic material or the sporadic collaboration among them incurs in a better evolution of all of them in an unison way. 

Some exponents of these methods can be found in the works \cite{cheng2017coevolutionary}, \cite{osaba2020coevolutionary} and \cite{osaba2020coeba}. All these three algorithms are multipopulation approaches, governed by separated Genetic Algorithms, Variable Neighborhood Search and Bat Algorithms, respectively. These three methods have demonstrated a promising performance, using a scheme in which each subpopulation is devoted to the solving of one single task. Furthermore, the genetic material exchange is materialized through the punctual migration of complete solutions among the multiple populations. The same trend is also adopted in \cite{zheng2019differential}, in which a MM method named as Differential Evolutionary Multitask Optimization is proposed, in which the knowledge sharing is conducted through the migration of individuals among populations. A similar philosophy is followed in the Multitasking Genetic Algorithm modeled in \cite{wu2020multitasking}, in which a population of solutions is created for each optimizing problem, and the knowledge sharing is realized at each iteration through the transference of different chromosomes among populations.

In the research conducted in \cite{feng2018evolutionary}, EM algorithm with explicit genetic transfer is presented. Also known as EM via autoencoding, or Explicit EM Algorithm, this method is comprised by as much independent populations as task being optimized. The knowledge sharing is materialized along the search through the injection of good solutions found by any of the subpopulations along their execution. For appropriately conducting this genetic transfer, a multiplication operation is used with a previously learned task mapping. Authors of this last work extend their research in \cite{feng2020explicit} by applying their EM via autoencoding to the well-known Capacitated Vehicle Routing Problem. Further evolution of this method is proposed in \cite{lin2020effective}, with an algorithm coined as EMT/ET. That enhanced technique explores a novel selection of transferred solutions, based on the dominance of that solutions over the optimizing problems. Additionally, in \cite{tang2019multipopulation} described a generalist multipopulation optimization scheme, based on similar concepts above described. Authors empirically demonstrate the efficiency of their scheme using a DE algorithm as base, giving rise to a so-called multipopulation multitask DE optimization. More concretely, this method capitalizes the sharing of information by sporadically creating overlapping populations. 

Also interesting is the work proposed in \cite{zhang2020multitasking}. In that work a specific instantiation of a multitasking genetic fuzzy system is presented and developed: a multitasking evolutionary optimization algorithm for Mamdani fuzzy systems with fully overlapping triangle membership functions (FOTMF-M-MTGFS). Further MM schemes can be found in \cite{liu2018surrogate,wang2021surrogate,zhang2021surrogate}, introducing surrogated-assisted mechanisms; or in \cite{chen2018fast}, which introduces a fast memetic algorithm.

Lastly, in \cite{xu2021multiobjective} a multi-objective multifactorial immune algorithm is proposed. That MM method works with different subpopulations, and bases the knowledge transfer on an explicit mechanism coined as Dimensional Information of Solutions (DIS). Thanks to this mechanism, subpopulations exchange their individuals selecting tasks with similar iteration trends.

\subsection{Explicit Knowledge Transfer Based Adaptive Solvers}
\label{sec:ExplicitAdaptive}

We finish this systematic review along the state of the art related to EM delving on the last category that can be found in the literature: explicit knowledge transfer based adaptive solvers. In this case, it is also interesting to mention that the methods than can be framed in this last category mainly embrace the above introduced MM philosophy.


In \cite{shang2019preliminary}, an interesting adaptive version of the above described Explicit EM Algorithm \cite{feng2018evolutionary} is proposed. Specifically, authors explore the use of the feedback gathered from the solutions transferred across tasks as guide for tasks selection. This feedback is updated along the search process, being able in this way to obtain the usefulness across tasks. An additional valuable algorithm is the novel EM algorithm with dynamic resource allocating strategy (MTO-DRA) introduced in \cite{gong2019evolutionary}. The adaptive mechanism considered in this EM method is similar to those presented in \cite{wen2017parting} or \cite{yao2020multiobjective}. Main novel ingredients of MTO-DRA in comparison with those similar methods is its multipopulation nature. More concretely, at each iteration, subpopulations are generated from the overall main population, each one fully devoted to the solving of one specific task. After this step, the resources are allocated to every subpopulation based on the index of improvements of tasks. This index is calculated online based on the performance feedback of previous generations. 

Authors in \cite{da2018curbing} propose an online similarity learning strategy, named as adaptive model-based transfer (AMT). For demonstrating its good performance, authors instantiate an EM algorithm, called AMT-\textit{enabled} EA. Main characteristic of the modeled AMT is its capability of dynamically learn and exploit the similarities across black-box optimization problems, minimizing negative transfers. This algorithm is further studied and enhanced in \cite{limnon} by means of online data-driven learning of non-linear mapping functions. 

Interesting is also the recent work proposed by Lim et al. in \cite{lim2021solution}. In that paper, a multiobjective probabilistic model-based transfer evolutionary optimization technique is proposed, endowed with a solution representation learning mechanism. More concretely, aligned solution representations are learned through spatial transformations. Thus, the technique is capable of tackling handle mismatches in search space dimensionalities among solving tasks, and also of increasing the overlap between search distribution of tasks. It is also interesting to mention that algorithms proposed in both \cite{da2018curbing} and \cite{lim2021solution} count with a single population, not being classifiable as MFO nor MM. Additional alternative adaptive EM schemes can be found in \cite{feng2017autoencoding,wang2021solving,gupta2019multitask,wei2021study}.

Throughout this systematic literature review section, we have conducted a deep analysis on the efforts made so far in Evolutionary Multitask Optimization field. In the next section \ref{sec:metho}, we further analyze and discuss on the common methodological trends observed in the literature. This methodological overview should also serve as guidance for the upcoming challenges related to this promising field.

\section{Current Methodological Trends in Evolutionary Multitask Optimization}
\label{sec:metho}

We provide in this section a methodological overview of the current state of EM research field. The studies already published in this area have been really abundant up to date, giving rise to a significant amount of techniques which share common practices, mechanisms and resources. The main reason of the existence of these different research trends is because they are dedicated to tackle some latent implementation challenges that should be addressed when dealing with EM environments. Thus, main goal of this section is to briefly highlight the principal methodologies adopted by practitioners in the different phases of algorithmic development.
\begin{itemize}[leftmargin=*]
	\item \textit{How to design the unified search space}: One of the most important issues when facing EM environments is the way in which solutions are encoded. This is essential principally in approaches that fall inside MFO paradigm. The main challenge at this point is that wide and generalist encodings will fall into superficial representations of solutions, not concrete enough for scrutinize interesting regions of task-specific search spaces. On the contrary, very specific representations can make impossible the genetic sharing between tasks coming from different optimization problems. In this sense, it should be taken into account that, depending on the type of EM technique to be implemented, it is possible that generated individuals are evaluated in different tasks throughout the whole search process. This is common in methods in which the sharing of knowledge is conducted by \textit{explicit transfer}. In other cases, although individuals are dedicated to the solving of an exclusive task, the existence of \textit{implicit transfer} procedures make essential that solutions devoted to the facing of different problems are capable of sharing knowledge with each other. This situation unveils the necessity of the existence of a unified search space, even more when tasks to solve are not completely related or belong to different typology of optimization problems. In the literature, many approaches for the efficient design of the unified search space can be found. If the tasks to solve are codified by continuous variables, the most used method for encoding individuals is the well-known random-keys representation \cite{bean1994genetic}, as can be seen in works such as \cite{gupta2016multiobjective,xiao2019multifactorial,zheng2019self,ong2016evolutionary,tang2019adaptive}. Furthermore, for discrete problems, two alternatives have been mainly followed by researchers: the transformation of the discrete search space to a continuous one through the random-keys representation, as mentioned in \cite{gupta2015multifactorial} and adopted in works as \cite{zhou2016evolutionary}; or the use of discrete search spaces such as the one introduced in \cite{yuan2016evolutionary} and used in works such as \cite{osaba2020mfcga,feng2020explicit}. Additional examples of encoding strategies can be found in the literature, but mainly constructed ad-hoc for a specific type of problems. Examples of this claim are the codification used in \cite{thanh2020efficient,thanh2018effective,thanh2020two} for solving clustered shortest path tree problems; or the one employed in \cite{osaba2020coevolutionary} for community detection over graphs. More recently, the use of neural auto-encoders has been proposed as a means to realize information transfer between tasks explicitly through the exchange of problem solutions, rather than delegating this exchange implicitly in the crossover operation over a unified search space \cite{feng2018evolutionary,feng2020explicit}.
	
	\item \textit{How to evolve the population(s) along the execution}: As mentioned before, Evolutionary Multitask Optimization refers to the design and implementation of multitasking solvers based on search procedures and operators drawn from Evolutionary Computation and Swarm Intelligence. Thus, as being population-based iterative methods, a crucial aspect that define this type of methods is the selection of the chromosomes that survive from one generation to another. In EM, several procedures have been proposed up to date, being the \textit{scalar fitness based selection} of MFEA the most often employed one. Specifically, \textit{scalar fitness based selection} is an elitist survivor function in which the best $\mathbf{P}$ individuals in terms of scalar fitness $\varphi^p$ among those in the current population and the newly produced offspring survive for the next generation. Another alternative strategy is the coined as \textit{local improvement	selection}, by which the newly generated solutions can only substitute their direct parent if they improve it. This strategy is followed in methods such as MFCGA \cite{osaba2020multifactorial,osaba2020transferability}, or those based on PSO or DE, as in \cite{xiao2019multifactorial,yu2019multifactorial}. On another vein, in most of MM schemes, the survivor selection is conducted within each subpopulation, following traditional evolutionary computation and/or swarm intelligence selection operators.
	
	Before proceeding further, at this point we critically pause at certain practices that have been lately noted in this area: the derivation of new EM approaches in which the novelty exclusively resides in the use of new search operators, without any other research contribution to the EM area whatsoever. Despite the interest that the exploration of new meta-heuristic operators may awake in the community, this uprising corpus of literature should be appraised with care. As any other subarea of Evolutionary Computation and Swarm Intelligence, many voices are claiming to cease research efforts towards metaphor-based studies that lack any algorithmic novelty when compared to well-established meta-heuristic approaches \cite{del2019bio}. Given its relative infancy, EM research should escape from these poor practices, and should focus strictly on algorithmic designs that connect closely to the core functionalities expected for an EM approach: new knowledge transfer mechanisms, unified solution encoding strategies, negative transfer avoidance over the search, and other parts alike.
	
	\item \textit{How to share knowledge between tasks}: The effective genetic material transfer is arguably the most important factor for EM methods to work in an efficient way. This specific procedure is what makes a multitasking technique to be superior to classical solving metaheuristics and schemes. In any case, the design of adequate knowledge sharing mechanisms is not trivial, and it usually depends on several issues, such as the encoding strategy employed, or the nature of the problems being solved. The main challenge on this point is twofold: i) to share as much as valuable knowledge among tasks and with an acceptable frequency, and ii) to define which is the genetic material that should be shared among optimizing tasks. Following these principles, the transference of knowledge can be capitalized following several directions. Probably the most used strategy, having shown great performance so far, is the generation of new individuals using genetic material coming from solutions with different skill task. Example of this specific trend is the well-known \textit{assortative mating}, which can be materialized through i) a common crossover operation as in MFEA, MFEA-II and many other MFO techniques \cite{gupta2015multifactorial,bali2019multifactorial,osaba2020mfcga}; ii) based on mutation strategies as in DE inspired techniques \cite{wang2019multiobjective,zhou2019towards,tang2020regularized}; or iii) the velocity based movements of PSO inspired methods \cite{feng2017empirical,xiao2019multifactorial,tang2019adaptive}. Another commonly used mechanism for conducting intra-task knowledge transfer is the one used by MM methods, in which multiple populations coexists, each one devoted to the resolution of one specific task \cite{cheng2017coevolutionary,osaba2020coeba,wu2020multitasking,chen2018fast,osaba2020coevolutionary}. In that cases, the knowledge sharing is conducted mainly by migrating solutions among subpopulations, modifying in this way the optimizing task of individuals. Other less used genetic transfer scheme is the one based on a single layer denoising autoencoder, used in the coined as EM via autoencoder \cite{feng2018evolutionary,feng2020explicit}; the generation of temporary overlapping populations \cite{tang2019multipopulation}, or based on the archives of DE solvers \cite{jin2019study}.

	\item \textit{How to adapt the algorithm to negative transfer}: as has been seen in previous Section \ref{sec:ImplicitAdaptive} and \ref{sec:ExplicitStatic}, a common trend for overcoming the curse of \textit{negative transfer} is the design and implementation of adaptive mechanisms. The main motivation that inspires the development of this mechanism is also twofold: i) to share as much as valuable knowledge among synergistic tasks and ii) to avoid the inefficient transfer of genetic material among non-complementary tasks.	In this regard, several promising alternatives have been proposed in the literature up to now. More concretely, we could distinguish two methodological trend: i) \textit{soft negative transfer avoidance} mechanisms, which are devoted to discourage the knowledge sharing among non-related tasks, and ii) \textit{hard negative transfer avoidance} mechanisms, aiming at prohibiting the transfer of genetic material among non-compatible tasks. Arguably, the most common used soft mechanism is the on-line fine tuning of algorithm parameters. This is the focal point of the influential MFEA-II \cite{bali2019multifactorial} and its discrete and multi-objective variants \cite{osaba2020dmfea,bali2020cognizant}, for example. Additional examples of this trend can be found in \cite{binh2019multi,chen2017adaptive,tang2019adaptive}, using similar mechanisms as MFEA-II, or in \cite{zhou2020toward,osaba2020mfcga,wang2019multiobjective}, in which the adaptation is not in the parameters, but in the search operators used. Another common strategy is the resource allocation \cite{wen2017parting,gong2019evolutionary,yao2020multiobjective}. This mechanism is in charge of dynamically analyze the complementarities among tasks and allocating computational resources based on them. Another \textit{soft} strategies contemplate the reinitialization of algorithm structures such as populations \cite{osaba2020mfcga,chen2017adaptive} or the controlling of the amount of genetic material exchanged among solutions \cite{liaw2017evolutionary,liaw2019evolutionary}. As \textit{hard} mechanism, we can highlight the dynamic generation of conceptual groups based on the arisen synergies \cite{tang2018group,chen2019adaptive}.  In any case, it should be highlighted that \textit{hard} are more complex to implement, since they should be aware of the similarities among optimizing tasks (either in a preliminary way, obtaining them dynamically, or by studying the corresponding landscapes).
	
\end{itemize}
	
Despite these methodological trends, other great challenges and niches persist in the field. Some of these research opportunities are very closely linked to the trends discussed in this section, while others are devoted to finding new methodological approaches or the application of EM technique to new and more complex domains. We review these challenges in Section \ref{sec:whereweshouldgo}, along with an outline of several research opportunities that are bound to attract much of the activity of the related community in the coming years.

\section{Evolutionary Multitask Optimization: Challenges and Research Directions}
\label{sec:whereweshouldgo}

\begin{figure}[h!]
	\includegraphics[width=\textwidth]{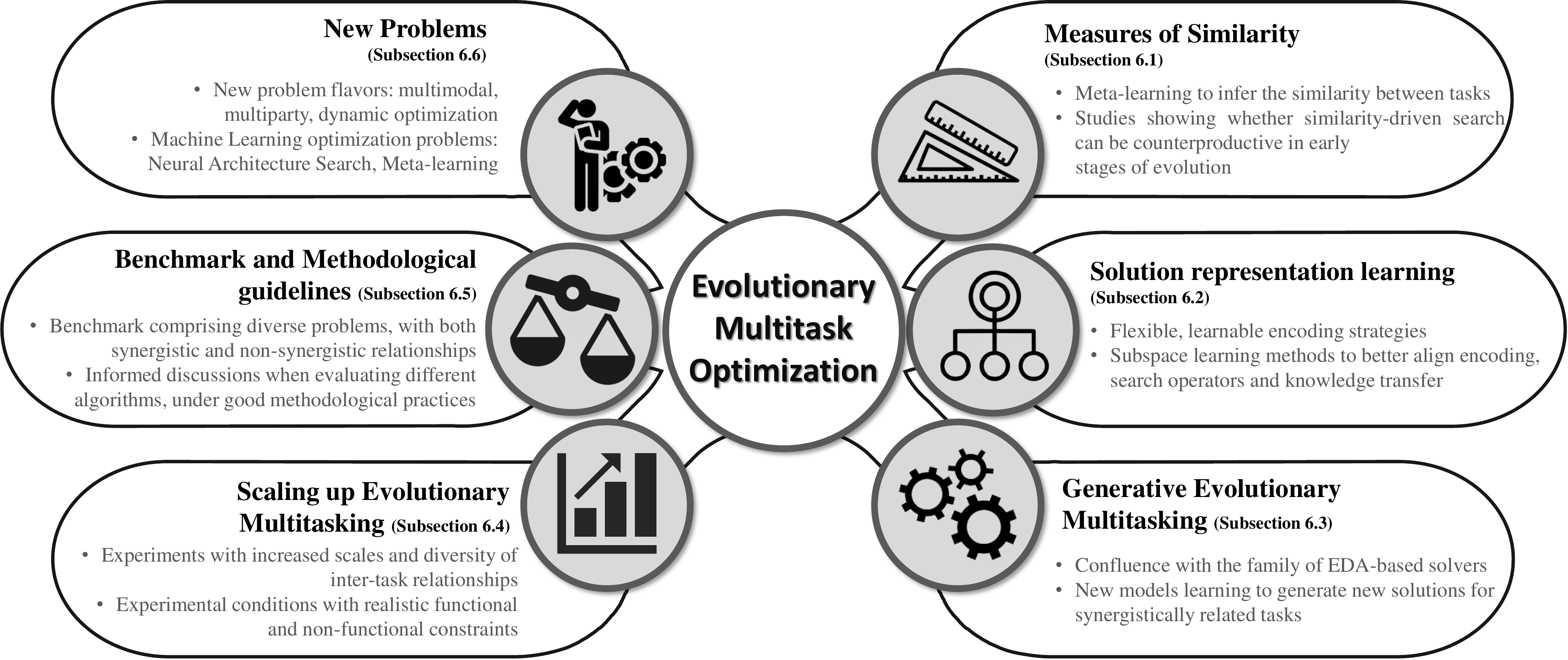}
	\caption{Conceptual diagram summarizing the identified set of challenges and research directions.}
	\label{fig:challenges}
\end{figure}

Considering the review of the activity so far discussed in preceding sections, there is little doubt that Evolutionary Multitasking has brought a fresh breeze to the community working on Evolutionary Computation and Swarm Intelligence. Advances so far in this area have been notable, exposing the benefits of embracing multitasking in optimization problems close to reality. However, the relative youth of this field has left several challenges and research niches still insufficiently addressed. In this section we enumerate a series of open research questions, and postulate research paths that can be followed to tackle them effectively in years to come. We complement each identified challenge by a brief explanation of its scope, relevance and alignment with current research efforts made in other fields, summarizing all this information in Figure \ref{fig:challenges} for a quick visual reference of contents:

\subsection{On Measures of Similarity between Tasks: Are they really needed?}

As discussed throughout the survey, so far the exchange of information between tasks has been done either implicitly or explicitly. In both cases, the similarity between tasks has been used to dictate which (and to a point, how) different individuals have been mated with each other, or to establish which tasks exchange explicit genotype information with each other. Notwithstanding this general usage pattern, an open question remains whether a priori assessment of the similarity between tasks is really needed in the context of Evolutionary Multitasking. Adaptive approaches such as MFEA-II- or ATMFCGA have exposed the capability of the evolutionary search process itself to elicit a progressively better estimation of the similarities between the problems being solved. However, there is no certainty whether this estimation of the similarity between tasks effectively avoids counter-synergies among them all along the process, particularly in early evolutionary stages. The availability of a priori information on how tasks relate to each other, by any means, should be exploited from the very beginning of the EM approach for the initial evolution to be informed properly.

Departing from this last intuition, we foresee that further efforts should be invested on advanced methods to estimate the similarity between optimization tasks without actually solving them. Clearly, a well-behaved measure of similarity between optimization tasks should roughly depend on the closeness of their optimal solutions. However, it is important to note that in the context of EM, the similarity between optimization tasks has no unique definition, and depends on the search and transfer operators being deployed. For instance, small differences in the solutions of two tasks can be amplified if the encoding strategy is not designed suitably, eventually leading to a counterproductive exchange of knowledge. This possibility is often overseen in the literature in favor of the design of unified representational strategies for all tasks under consideration. Conversely, given certain search operators, the tasks can be claimed to be related/similar to each other only in the context of the encoding strategy and operators in use, and provided that multitasking leads to faster convergence than in the case of isolated problem solving. The same set of problems may lead to negative knowledge transfer if different operators are used.

We definitely advocate for further research in this direction. A first research direction to follow is the incorporation of meta-learning algorithms capable of inferring the similarity between pairs of tasks based on meta-features extracted from the problems (e.g. based on fitness landscapes or on solution space sampling). This similarity estimation should be also complemented by an encoding alignment between tasks that ensures maximally aligned individuals in multi-population EM approaches. For this latter purpose, non-linear methods from domain adaptation have been recently explored from the transfer optimization perspective \cite{limnon}, leaving a door open to the consideration of further ingredients from subspace learning.

\subsection{Solution Representation Learning: Blending together Encoding and Search Operators}

Grounded on the two schools of thought about how to face information transfer between tasks (explicit versus implicit), a further step should be taken towards finding not only solutions to the problems, but also representation of the solutions for each problem that are more efficient for conveying knowledge transfer among tasks. This resonates with the reflections made in the previous subsection, by which similarity is strongly subject to the set of operators and the encoding strategy in use. Indeed, knowledge exchange between tasks can be beneficial only under appropriate solution representations. For instance, in graph coloring/community detection problems, a permutation-invariant encoding approach has been noted to be of utmost necessity for implicit information transfer through crossover \cite{osaba2020coevolutionary}. Otherwise, information transfer cannot lead to better convergence, even if the networks to be clustered are rotated versions of the same network. 

Solution representation learning is therefore vital for effective knowledge transfers. This is an exceedingly important topic for future research in evolutionary multitasking: learning solution representations, either based on prior data or adaptively during the course of the search, to enhance positive transfers. This unleashes an interesting opportunity for \emph{learnable} encoding strategies, especially for those that can be evolved jointly with the solution itself (e.g. genetic programming). Otherwise, when allowed by the application domain where tasks are defined, tailored alignment methods or flexible encoding approaches should be utilized instead, always coupled tightly to the heuristic search operators in use.

\subsection{Learning to Search using Generative Evolutionary Multitasking}

Most EM approaches reported to date are based on sampling the space of possible solutions to the problem, without any attempt at learning the distribution of good solutions. In other words, the space of possible solutions is traversed by resorting to evolutionary and/or swarm operators, so that new solutions stem from the application of such operators to one or multiple populations of individuals. An additional degree of intelligence in how the space is sampled could be achieved by creating synthetic solutions along the search that reinforce and push forward the convergence of synergistically related tasks. If two tasks were found to be related to each other during the search, a generative machine learning model could progressively learn the distribution of \emph{good} solutions for both problems. Once learned, this generative model could be queried over the search, replacing (fully or partly) the application of evolutionary operators. As a result, synthetic solutions that are potentially good for related tasks could be produced and fed to the population, ultimately accelerating the convergence. 

The adoption of latent generative models already underlies beneath renowned EM methods, such as the probability mixture models used in MFEA-II to model the relationships between tasks. It is our belief that a profitable research path for EM remains in the long history of EDA algorithms, which address the concept of generative modeling and sampling for single-task optimization. It will be a matter of time when the EDA and EM realms collide together to span a new generation of intelligent multitask solvers, not only producing solutions, but also distributions that can be exported for other EM setups comprising task instances of similar kind.

\subsection{Scaling up Evolutionary Multitasking: is it just a Matter of Search Algorithms?}

In reduced experimentation setups the use of EM methods has been shown to yield benefits in terms of convergence with respect to single-task optimization. However, when scaling up EM environments to realistic levels in terms of the number and diversity of tasks, these observed benefits can be turned down due to several reasons. To begin with, the computational resources required to scale up the search nicely with the number of tasks can become not affordable if the search over all tasks is to be made in a centralized fashion. An opportunity arises at this point for multipopulation schemes at the expense of multifactorial approaches, as they naively allow for decentralized implementations of the search and thereby, a more balanced share of the computational cost among stakeholders. This alternative, however, would come along with other aspects to be considered, such as the selection of a synchronous/asynchronous knowledge transfer policy or the reliability of the fitness evaluation made locally, among other issues noted in the field of distributed evolutionary computation \cite{gong2015distributed}.

Even if the above matters become eventually solved by the advance in research, we definitely need to formulate an additional question: when and where can it be beneficial to solve thousands (potentially, millions) of tasks at once? Is there any realistic setup comprising these scales, in which several problems are related to each other so that this synergy leads to quantifiable performance gains? It is an undeniable truth that so far, experimental setups utilized in the community working in EM have been restricted to a few selected problem instances as per their relationship known beforehand. Non-functional aspects inherent to a distributed setup are, therefore, left aside in favor of a more focused pursuit towards algorithmic advantages. In real settings, however, other aspects should be under study, which could imply modifications at the core of EM approaches. 

Among such aspects, promising paths have been lately traversed in what regards to the scalability of EM approaches with respect to the number and complexity of tasks \cite{shakeri2020scalable}. In this work another important issue arising when deploying EM approaches in large-scale settings was identified: the efficiency of learning over the search how to transfer knowledge among a sparsity of related tasks. This work should stimulate increasing attention of the community in the need for a profound redesign of existing EM mechanisms when the practical setup in which they are evaluated comprises a more realistic mixture of related and unrelated tasks.

Privacy guarantees as those sought in affine modeling fields (e.g. federated learning, differential privacy, homomorphic computation) could be another aspect of relevance when scaling up EM frameworks. Delving into this matter, \emph{federated optimization} would aim to evolve jointly different distributed tasks, without each task revealing each other their actual best solutions. A possible approach would be to define an encrypted unified search space, so that only each task could decipher its corresponding solution. The challenge in this direction is how to include this encryption functionality without jeopardizing the transfer of knowledge via implicit genetic transfer, or hindering the overall multitasking search efficiency.

\subsection{On the Need for Diverse Benchmarks and Methodological Experimentation Guidelines}

One aspect of EM research that has been put to question is the quality of experimental benchmarks designed to assess the performance of every proposed approach. Most contributions to date have traditionally considered experimental setups comprising a few tasks, at their best belonging to 4-5 problem formulations over which inter-task similarities and synergies can be analyzed and discussed. Despite recent efforts in the heat of competitions held in frontline conferences\footnote{Competitions on Evolutionary Multi-task Optimization held at IEEE Congress on Evolutionary Computation (CEC'2017 to CEC'2021) and Genetic and Evolutionary Computation Conference (GECCO'2020).}, common methodological guidelines and benchmarking tools are still to be agreed and widely adopted in prospective contributions. Otherwise, there will be no clear grounds to ensure the fair evaluation, replication and comparison of new advances in the field.

Therefore, new benchmarks, quantitative metrics and methodological guidelines should be proposed, discussed and embraced by researchers working in EM. On the one hand, scores should relate to the quality of the produced solutions for the tasks under consideration, as well as the computational efficiency of the joint search, the gains with respect to single-task optimization, and the amount of positive/negative transfer episodes registered for every task pair. Finally, methodologically speaking all aspects that impact on the obtained simulation outcomes should be reported, especially those that are often overseen when describing the experimental setup (e.g. parameter tuning of all solvers under comparison, the imposed convergence criterion, and a solid justification why the selected parameters make the comparison fair among solvers). Furthermore, the usual discussion among approaches held on the basis of global performance statistics (e.g. average fitness per task) should be informed with additional null hypothesis tests \cite{derrac2011practical} and/or a Bayesian characterization of the obtained results \cite{carrasco2020recent} to guarantee the statistical significance of the gaps claimed to exist among different approaches. All in all, a major push towards crystal clear comparisons in all measurable aspects of multitasking.

\subsection{New Problems in Evolutionary Multitasking: Multimodality, Metalearning and Beyond}

In the last couple of years, a growing corpus of literature has addressed EM for tasks that go beyond real-valued single-objective optimization problems. This is the case of permutation-based combinatorial and multi-objective optimization, which have been tackled with EM approaches that incorporate algorithmic ingredients suited to deal with these problems. However, other flavors have been addressed more scarcely to date. This includes multimodal optimization, where synergies emerge as per the number and inter-distance between global optima shared by the tasks; dynamic optimization, particularly the case when changes undergone by two tasks occur in the same direction yet at different instants over time; or multiparty optimization, where several stakeholders participate, sharing part of the objectives and/or the solutions of their related Pareto front approximations.  

An application domain that deserves a separate mention at this point is the confluence between Machine Learning (ML) and EM. Indeed, many learning algorithms can be formulated as an optimization problem (e.g. loss minimization in Deep Neural Networks), therefore unleashing an opportunity for undertaking setups consisting of several interrelated ML problems with EM approaches. For instance, it has been widely postulated that Evolutionary Computation and Swarm Intelligence solvers can be used as an scalable replacement for optimization problems related to Deep Neural Networks \cite{martinez2020lights,salimans2017evolution,iba2020deep}. Initial explorations have exposed that EM can be used in multitask reinforcement learning environments to jointly train the neural models and exploit synergies between them \cite{martinez2020simultaneously}. However, other avenues at this crossroads are worth to be explored further, such as neural architecture search, where the joint evolution can serve as a mutual guidance for avoiding regions representing underperforming network configurations; and meta-learning, where the paradigm resides in how to optimize models that can perform well in unseen tasks. ML problems will surely unfold an interesting playground for multitask optimization in the near future.

\section{Concluding Remarks and Outlook}
\label{sec:conc}

This overview has elaborated on the research area known as Evolutionary Multitask Optimization. Framed within the wider Transfer Optimization field, the main goal of this incipient paradigm is to exploit the knowledge learned throughout the optimization of one problem towards addressing other related or unrelated problems, so that they are solved more effectively than in isolation. The youth of this area clashes with the relatively high amount of contributions reported by the community to date. Consequently, our study aims at analyzing the past, present and future of this area, emphasizing on methodological patterns, practices and concepts followed by the community. To this end, we have first delved into the essentials of EM, establishing mathematical grounds that allow discerning the aforementioned methodological patterns in subsequent discussions. Furthermore, a clear distinction between multitask optimization, multiobjective optimization and multitask learning has been done. Departing from these prior definitions, we have performed a systematic review of the literature related to EM, focusing on remarkable studies published in the last few years, and establishing a landmark taxonomy that allows the audience to easily understand the algorithmic choices mostly embraced in the reviewed bibliography. Specifically, our study has informed about the prominent role of multifactorial optimization and multipopulation multitasking approaches. We have also conducted a methodological analysis of the different phases followed when designing EM solvers. Finally, we have built upon our critical literature analysis to initiate a discussion around research niches and challenges that remain insufficiently addressed to date. On a prescriptive note, each of such identified challenges has been associated with several possible research directions, which should inspire efforts in years to come.

Exchange of knowledge between researchers should be encouraged via a common space of understanding in which to unify their views, synchronize their research agendas and push energetically their efforts towards valuable advances in the field. The ultimate purpose of this work is to promote synergistic knowledge transfer between researchers working in Evolutionary Multitasking, much in line with what is sought in multitasking itself. We also hope that this material enshrines as a suggestive point of reference for newcomers and practitioners interested in a smooth landing at this fascinating area.

\section*{Acknowledgements}

The authors would like to thank the Basque Government for its funding support through the ELKARTEK program (3KIA project, KK-2020/00049) and the consolidated research group MATHMODE (ref. T1294-19).

\bibliography{biblio}

\section*{Author biographies}
\noindent \textbf{Eneko Osaba} works at TECNALIA as researcher in the ICT/OPTIMA area. He received the B.S. and M.S. degrees in computer sciences from the University of Deusto, Spain, in 2010 and 2011, respectively. He obtained his Ph.D. degree on Artificial Intelligence in 2015 in the same university, being the recipient of a Basque Government doctoral grant. Throughout his career, he has participated in the proposal, development and justification of more than 25 local and european research projects. Additionally, Eneko has also participated in the publication of more than 125 scientific papers (including more than 25 Q1). Eneko has obtained three national accreditations from the National Agency for Quality Assessment and Accreditation of Spain, ANECA: PhD lecturer, PhD lecturer of private university and PhD assistant lecturer. Additionally, in 2016-2017, he has participated as coordinator and teacher of the Research Seminar on Academic Writing in Engineering, and as teacher in the Research Seminar on the Research Career in Engineering, both of them part of the PhD Program in Engineering for the Information Society and Sustainable Development of the University of Deusto. He has performed several stays in universities of United Kingdom (Middlesex University), Italy (Universitá Politecnica delle Merche) and Malta (University of Malta). Eneko has served as member of the program committee in more than 50 international conferences. Furthermore, he has participated in organizing activities in more than 10 international conferences. Besides this, he is member of the editorial board of International Journal of Artificial Intelligence, Data in Brief and Journal of Advanced Transportation, and he has acted as guess editor in journals such as Journal of Computational Science, Neurocomputing, Logic Journal of IGPL, Swarm and Evolutionary Computation, Advances in Mechanical Engineering journal and IEEE ITS Magazine. In his research profile it can be found a 19 H-index with more than 1500 cites in Google Scholar. Additionally, Eneko was an Individual Ambassador for ORCID along 2017-2018. Finally, he has nine intellectual property registers, granted by the Basque Government, and he has two European patents under review.
\subsection*{  } 
\noindent \textbf{Aritz D. Martinez} studied Telecommunications Engineering at the University of the Basque Country (UPV/EHU, Spain), he obtained his M.Sc. degree in Computational Engineering and Intelligent  Systems from this university in 2018 and now he is working towards his PhD at Tecnalia Research and Innovation. His research interests are focused on heuristic techniques for optimization, parallel  computing, delay tolerant networking, Deep Learning, Reinforcement Learning and Federated Machine Learning, among others.
\subsection*{  } 
\noindent \textbf{Javier Del Ser} received his first PhD degree (cum laude) in Electrical Engineering from the University of Navarra (Spain) in 2006, and a second PhD degree (cum laude, extraordinary PhD prize) in Computational Intelligence from the University of Alcala (Spain) in 2013. He is currently a Research Professor in Artificial Intelligence and leading scientist of the OPTIMA (Optimization, Modeling and Analytics) research area at TECNALIA, Spain. He is also an adjunct professor at the University of the Basque Country (UPV/EHU), and an invited research fellow at the Basque Center for Applied Mathematics (BCAM), both also located in Spain. His research interests are in the design of Artificial Intelligence methods for data modeling and optimization applied to problems emerging from a manifold of domains, including manufacturing industries, intelligent transportation systems, logistics, health, telecommunications and energy, among others. In these fields he has published more than 380 scientific articles, co-supervised 10 Ph.D. theses, edited 7 books, co-authored 9 patents and participated/led more than 40 research projects. He is an Associate Editor of tier-one journals from areas related to Artificial Intelligence, such as Information Fusion, Swarm and Evolutionary Computation, Cognitive Computation and IEEE Transactions on Emerging Topics in Artificial Intelligence.

\end{document}